\documentclass[runningheads]{llncs}
\usepackage{graphicx}

\usepackage{comment}
\usepackage{amsmath,amssymb} 
\usepackage{color}
\usepackage[super]{nth}
\usepackage{animate}
\usepackage{nth}
\usepackage{orcidlink} 
\usepackage{float}
\usepackage{tikz} %
 
\usepackage[accsupp]{axessibility}  

\usepackage{pifont}
\newcommand{\cmark}{\ding{51}}%
\newcommand{\xmark}{\ding{55}}%
\usepackage{booktabs}
\usepackage[capitalize]{cleveref}
\usepackage{bm}
\crefname{section}{Sec.}{Secs.}
\Crefname{section}{Section}{Sections}
\Crefname{table}{Table}{Tables}
\crefname{table}{Tab.}{Tabs.}
\usepackage{amsmath}

\DeclareMathOperator*{\argmin}{arg\,min}

\usepackage{xspace}
\makeatletter
\DeclareRobustCommand\onedot{\futurelet\@let@token\@onedot}
\def\@onedot{\ifx\@let@token.\else.\null\fi\xspace}

\def\eg{\emph{e.g}\onedot} 
\def\ie{\emph{i.e}\onedot} 
\def\cf{\emph{c.f}\onedot} 
 
\def\wrt{w.r.t\onedot} 

\def\etal{\emph{et al}\onedot}
\makeatother

\newcommand{\beginsupplement}{%
        \setcounter{table}{0}
        \renewcommand{\thetable}{S\arabic{table}}%
        \setcounter{figure}{0}
        \renewcommand{\thefigure}{S\arabic{figure}}%
     }

\makeatletter
\newcommand\refwithdefault[2]{%
  \@ifundefined{r@#1}{%
    #2%
  }{%
    \ref{#1}%
  }%
}
\makeatother

\begin{document}
\pagestyle{headings}
\mainmatter
\def\ECCVSubNumber{7092}  

\title{PREF: Predictability Regularized Neural Motion Fields}

\titlerunning{PREF: Predictability Regularized Neural Motion Fields}
%
\author{Liangchen Song\inst{12}\orcidlink{0000-0002-8366-5088} \and
Xuan Gong\inst{12}\orcidlink{0000-0001-8303-633X} \and 
Benjamin Planche\inst{2}\orcidlink{0000-0002-6110-6437} \and
Meng Zheng\inst{2}\orcidlink{0000-0002-6677-2017} \and \\
David Doermann\inst{1}\orcidlink{0000-0003-1639-4561} \and
Junsong Yuan\inst{1}\orcidlink{0000-0002-7901-8793} \and
Terrence Chen\inst{2} \and 
Ziyan Wu\inst{2}\orcidlink{0000-0002-9774-7770}}
\authorrunning{L. Song et al.}
%
\institute{University at Buffalo, Buffalo NY, USA \and
United Imaging Intelligence, Cambridge MA, USA \\
\email{\{first.last\}@uii-ai.com}}
\maketitle

\begin{abstract}
Knowing the 3D motions in a dynamic scene is  essential to many vision applications. Recent progress is mainly focused on estimating the activity of some specific elements like humans. In this paper, we leverage a neural motion field for estimating the motion of all points in a multiview setting. Modeling the motion from a dynamic scene with multiview data is challenging due to the ambiguities in points of similar color and points with time-varying color. We propose to regularize the estimated motion to be predictable. If the motion from previous frames is known, then the motion in the near future should be predictable. Therefore, we introduce a predictability regularization by first conditioning the estimated motion on latent embeddings, then by adopting a predictor network to enforce predictability on the embeddings. The proposed framework PREF (\textbf{P}redictability \textbf{RE}gularized \textbf{F}ields) achieves on par or better results than state-of-the-art neural motion field-based dynamic scene representation methods while requiring no prior knowledge of the scene.
\end{abstract}


\begin{figure}[t]
    \centering
    \begin{animateinline}[controls,buttonsize=.7em,autoplay,loop,palindrome,poster=last,width=1.\textwidth]{5}
        \centering
        \multiframe{4}{i=0+1}{%
            \begin{tabular}{cc}
            (a) Topologically varying scene & 
            (b) Physical motion for all points
            \\
            \includegraphics[width=.5\linewidth]{fig/intro/seq1/img\i} & 
            \includegraphics[width=.5\linewidth]{fig/intro/seq2/img\i}
            \\
            \includegraphics[width=.5\linewidth]{fig/intro/seq1/plot\i} &
            \includegraphics[width=.5\linewidth]{fig/intro/seq2/plot\i}
            \\
            \multicolumn{2}{p{\textwidth}}{\scriptsize{To see the animation, please view the document with compatible software, \eg,  \textit{Adobe Acrobat} or \textit{KDE Okular}; otherwise, the animation is also provided as a separate file in supplementary material.}}
            \end{tabular}
        }
    \end{animateinline}
    \caption{Our method can handle topologically varying scenes and estimate physical motion for all points in the space. \textit{Topologically varying} means that the topology of the scene can change, such as a new person entering the scene in (a). All points in the space are tracked, such as the ball in (b). Only the sequence of images to be analyzed is used and no prior knowledge is required in our framework. 
    }
    \label{fig:intro}
\end{figure}

    

\section{Introduction}
Estimating motion in dynamic scenes is a fundamental and long-standing  problem in computer vision \cite{huang1981image}. 
Most of the existing 3D motion estimation works are concerned with specific objects like humans \cite{reddy2021tessetrack}. Still, knowing the 3D motion of all objects in a dynamic scene can be of great benefit to a number of vision applications like robot path planning \cite{chung2018survey}.
Tracking all points in the space with only multiview data is obviously challenging, however, neural fields is a hot topic that has emerged recently \cite{xie2021neural}, bringing hope to breakthroughs for this problem.

Neural fields, also known as coordinate-based neural networks, have demonstrated great potential in dynamic 3D scene reconstruction from multiview data \cite{TewariFTSLSMSSN20,xie2021neural}. 
Coordinate-based representations not only naturally support fine-grained modeling of the motion for points in space, but also require no prior knowledge about the geometry and track all points in space. In this paper, we address the problem of estimating 3D motion from multiview image sequences, for general scenes and for all points in the space (\cref{fig:intro}).

Despite recent progress on neural fields-based dynamic scene representation (\eg, \cite{lombardi2019neural,gafni2020dynamic,Park_2021_ICCV,pumarola2020d,li2020neural,xian2020space,Tretschk_2021_ICCV,li2021neural,park2021hypernerf,ZhangLYZZWZXY21,wang2021neural,du2021nerflow,fang2022fast}), estimating 3D motion from multiview data remains challenging for the following reasons. First, motion ambiguity exists among points with the same 
color, so
one cannot confidently track interchangeable points on non-rigid surfaces from visual observations alone (\cf possibility of position swapping).
Second, the color of any point may change over time. For example,  spatially or temporally varying lighting conditions can blur the notion of a point's identity over time.

In this paper, we propose to \emph{regularize the estimated motion to be predictable}  to address the aforementioned ambiguity issues. 
The key insight behind motion predictability is that underlying motion patterns exist in a dynamic real-world scene. Chaotic motions (\eg, position swapping for similarly-colored points) are not predictable and should be penalized.
In our work, the motion in a scene is ``\emph{implicitly}'' regularized by enforcing predictability, which is intrinsically different from explicitly designed regularizing terms, such as elastic regularization \cite{Park_2021_ICCV} and as-rigid-as-possible regularization \cite{Tretschk_2021_ICCV}.
State-of-the-art solutions use combinations of space-time radiance neural fields and neural motion fields to model dynamic scenes, optimizing these fields jointly over a set of visual observations in a self-supervised manner by comparing predicted images to actual observations. But vision-based supervision alone typically results in noisy and poorly disentangled motion fields, \cf aforementioned ambiguities. Therefore, some recent works use data-driven priors like depth \cite{xian2020space} and 2D optical flow \cite{li2020neural} as a regularization.
In contrast, we propose to improve motion field optimization through predictability-based regularization. Instead of learning a motion field $M$ that maps each 3D position $\mathbf{p}$ and timestep $t$ to a deformation vector $\Delta_{t \rightarrow t+\delta t}\mathbf{p}$,
we  condition the motion field on a predictable embedding of the motion for queried time (noted $\bm{\omega}_{t \rightarrow t+\delta t}$), \ie, $\Delta_{t \rightarrow t+\delta t}\mathbf{p} = M(\mathbf{p}, \bm{\omega}_{t \rightarrow t+\delta t})$. These motion embeddings are either directly optimized jointly with the space-time field over observations, or are inferred by a predictor function $P$ that takes a set of past embeddings and infers the next motion embedding. 
During scene optimization, we enforce each motion embedding regressed from the observations to be predictable by our model $P$. Therefore we promote the encoding of underlying motion patterns and penalize chaotic and unlikely-realistic deformations.
In summary, our contributions are as follows:
\begin{itemize}
    \item We propose to leverage predictability as a prior \wrt the motion in a dynamic scene. Predictability regularization implicitly penalizes chaotic motion estimation and can help solve the ambiguity of motion.
    \item We condition point motions on embedding vectors and design a predictor on the embedding space to enforce motion predictability.
    \item We demonstrate the benefits of the resulting additional supervision (predictability regularization) on motion learning through a variety of qualitative and quantitative evaluations.
    \item We provide insights into how the proposed framework can be leveraged for motion prediction as a by-product.
\end{itemize}


\section{Related Work}
\paragraph{\textbf{Neural fields.}}
A neural field is a field that is parameterized fully or in part by a neural network \cite{xie2021neural,tensorf}. 
Neural fields are widely used for implicitly encoding the geometry of a scene, such as occupancy \cite{mescheder2019occupancy} and distance function \cite{park2019deepsdf,chibane2020neural}. Our method is built on the milestone work NeRF \cite{mildenhall2020nerf}, in which the radiance and density are encoded in neural fields. NeRF led to a series of breakthroughs in the fields of 3D scene understanding and rendering, such as relighting \cite{boss2021nerd,srinivasan2021nerv,boss2021neuralpil}, human face and body capture \cite{hong2021headnerf,Noguchi_2021_ICCV,peng2021neural,Peng_2021_ICCV,su2021anerf,liu2021neural}, and city-scale reconstruction \cite{tancik2022blocknerf,xiangli2021citynerf,turki2021mega,rematas2021urf}. A recent method also named PREF \cite{Huang2022PREF} is developed for compact neural signal modeling.

\paragraph{\textbf{Motion estimation and 4D reconstruction.}}
Large-scale learning-based motion estimation from multiview data achieved impressive performance \cite{li2017robust,reddy2021tessetrack}, but most methods are constrained to tracking some specific objects such as humans \cite{reddy2021tessetrack}. In this paper, we are concerned with estimating the motion of all points without access to any annotations, which is related to the 4D reconstruction problem where motion is usually estimated. 
Some methods have been developed with known geometry information such as depth or point cloud. DynamicFusion \cite{newcombe2015dynamicfusion}, Schmidt \etal \cite{schmidt2015dart}, Bozic \etal \cite{bozic2020neural}, and Yoon \etal \cite{yoon2020novel} estimate motion from videos with depth. OFlow \cite{niemeyer2019occupancy} and ShapeFlow \cite{jiang2020shapeflow} infer a deformation flow field with the knowledge of occupancy. 
More recently, motivated by the success of NeRF, a number of methods have been designed to reconstruct 4D scenes as well as motion directly from multiview data, which can be acquired from a multi-camera system or a single moving camera. 
D-NeRF \cite{pumarola2020d}, Nerfies \cite{Park_2021_ICCV} and NR-NeRF \cite{Tretschk_2021_ICCV} set a canonical frame and align dynamic points to it. 
DCT-NeRF \cite{wang2021neural} proposes to track the trajectory of a point along all sequences. NSFF \cite{li2020neural}, VideoNeRF \cite{xian2020space}, and NeRFlow \cite{du2021nerflow} propose to represent the dynamic scene with a 4D space-time field, thus able to handle topologically varying scenes. The 4D fields are under-determined, and precomputed data-driven priors are usually needed to achieve good performance. HyperNeRF \cite{park2021hypernerf} proposes to align frames towards a hyperspace for topologically varying scenes and achieves state-of-the-art performance without the need of data-driven priors. These methods are able to render visually appealing images for novel views and time, 
yet their performance on 3D motion estimation has room for improvements. 

\paragraph{\textbf{Scene flow estimation.}}
3D motion field is also known as dense scene flow \cite{menze2015object,mittal2020just,zhai2021optical}. Vedula \etal \cite{vedula1999three} introduced the concept and demonstrated a framework for acquiring dense, non-rigid scene flow from optical flow. Basha \etal \cite{basha2013multi} proposed a 3D point cloud parameterization of the 3D structure and scene flow with calibrated multi-view videos. Vogel \etal \cite{vogel2013piecewise} suggested to represent the dynamic 3D scene by a collection of planar, rigidly moving, local segments. More recently, Yang \etal \cite{yang2021learning} proposed a framework adopting 3D rigid transformations for analyzing background segmentation and rigidly moving objects. 

\paragraph{\textbf{Predictability.}}
The study of the predictability of time series data dates back to \cite{box1977canonical,pena1987identifying}, in which predictability is interpreted as the ability to be decomposed into lower-dimensional components. The idea of extracting principal components as predictability is adopted for blind source separation in \cite{stone2001blind}. Differential entropy is used for measuring predictability in \cite{goerg2013forecastable}. Our method shares a similar motivation as the above methods in terms of discovering low-rank structures, while predictability in our method is not explicitly defined but implicitly introduced through a predictor network. 


\section{Preliminaries}

Our method is built upon the NeRF framework \cite{mildenhall2020nerf} and is inspired by recent progresses \wrt dynamic scenes \cite{xian2020space,li2020neural}. 
For each 3D point $\mathbf{p}=(x,y,z)$ in the considered space, we represent its volume density by $\bm{\sigma}(\mathbf{p})$, and its color from a viewing direction $\mathbf{d}$ by $\mathbf{c}(\mathbf{p},\mathbf{d})$.
In NeRF, these two attributes are defined as the output of a continuous function $F$ modeled by a neural network, \ie, $(\mathbf{c},\bm{\sigma})=F(\mathbf{p},\mathbf{d})$.
This neural field can be queried to render images of the represented scene through volume rendering.
For each camera ray $\mathbf{r}$ defined by its optical origin $\mathbf{o}$ and direction $\mathbf{d}$ intersecting a pixel, we compute the color $\mathbf{C}(\mathbf{r})$ of said pixel by sampling points along the ray, \ie, sampling $\mathbf{p}_i=\mathbf{o}+i\mathbf{d}$; then querying and accumulating their attributes according to $F$. Overall, the expected color $\mathbf{C}(\mathbf{r})$ of the ray $\mathbf{r}$ is:
\begin{equation}\label{eq:rendering}
\mathbf{C}(\mathbf{r})=\int_{i_n}^{i_f} e^{-\int_{i_n}^i\bm{\sigma}(\mathbf{p}_j) dj}\bm{\sigma}\big(\mathbf{p}_i\big)\mathbf{c}\big(\mathbf{p}_i,\mathbf{d}\big)di,
\end{equation}
where $i_n, i_f$ are near and far bounds. The integration in \cref{eq:rendering} is numerically approximated by summing up a set of points on the ray. 

For dynamic scenes, existing solutions can be roughly categorized into two groups. Either methods model the motion and radiance with two distinct fields \cite{pumarola2020d,Park_2021_ICCV}, or they are regularizing the motion from a space-time field \cite{li2020neural,xian2020space,du2021nerflow}. In the former solutions, the color of a point $\mathbf{p}$ at time $t$ is represented by $F_k(M(\mathbf{p}, t),\mathbf{d})$, where $F_k$ represents the $k$th canonical time-invariant space and $M$ is a learned neural motion field defining the motion $\Delta\mathbf{p}$ of any point $\mathbf{p}$ at time $t$ \wrt to their position in the canonical space. 
Our method falls into the latter category, in which each point in the dynamic scene is represented by a space-time field $F(\mathbf{p}, \mathbf{d}, t)$. 
Unlike canonical space-based methods, for the space-time field we need to specify the frame of $F$ when joint training with a motion field $M$.
We opt for space-time field rather than canonical-space one for two reasons. 
First, we presume that underlying patterns exist for the motion of a certain time range. 
So canonical-frame-based motion estimation frameworks are not suitable, since their motions are from the predefined canonical frame to another, whereas we need the motion between a certain range of frames.
Second, space-time fields are more generic as they can handle non-existent geometry in the canonical frame (\eg, objects entering the scene mid-sequence). 
Note that for both categories, the scene fields are optimized jointly leveraging observation-based self-supervision, \ie, computing the image reconstruction loss for each time step $t$ as:
\begin{equation}\label{eq:rec}
\mathcal{L}_{\mathrm{rec}}=\sum_{\mathbf{r}}\| \mathbf{C}_{\mathrm{gt}}^t(\mathbf{r}) - \mathbf{C}_{\mathrm{}}^t(\mathbf{r}) \|_2^2,
\end{equation}
with $\mathbf{C}_{\mathrm{gt}}^t$ is the observed pixel color and $\mathbf{C}^t$ is the color rendered from  $F$ and $M$.

\section{Method}
\subsection{Overview}
\begin{figure}[t]
    \centering
    \includegraphics[width=1.0\columnwidth,trim={0 0 0 0},clip]{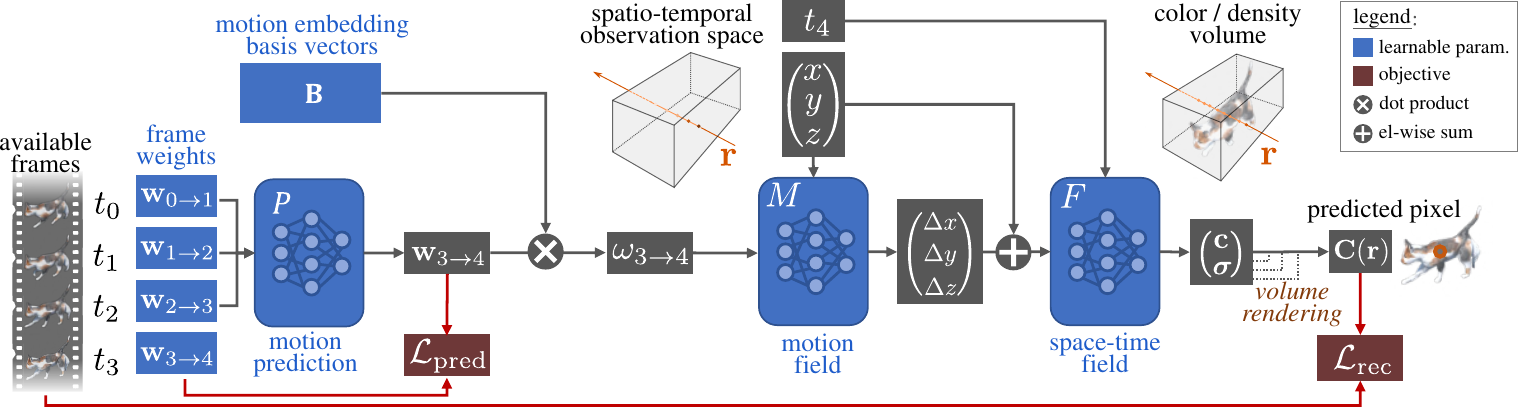}
    \caption{Overview of the proposed framework. Three networks are trained jointly: the space-time field, the motion field and the predictor. The space-time field returns color and occupancy for each point at a specific time. The motion field predicts the motion of a point based on a motion embedding vector. The predictor generates the future motion embedding based on previously observed embeddings.
    }
    \label{fig:overview}
\end{figure}

Our framework consists of three components: a neural space-time field $F$, a motion field $M$ and a motion predictor $P$. An overview of their interactions is presented in \cref{fig:overview}. In our framework and implementations, we do not model the viewing dependency effects with the space-time field, so the space-time field outputs the color and occupancy for each point $(x,y,z,t)$, whereas the motion field provides the motion of any point between two time steps, according to the space-time field. Let the motion of point $\mathbf{p}=(x,y,z)$ from time $t$ to $t+\delta t$ be $\Delta_{t\rightarrow t+\delta t}\mathbf{p}$, then for $\mathbf{p}$ at time $t$ we have:
\begin{equation}\label{eq:motion}
(\mathbf{c}_t,\bm{\sigma}_t)=F(\mathbf{p}+\Delta_{t\rightarrow t+\delta t}\mathbf{p},t+\delta t).
\end{equation}
The idea is that for a scene observed at time $t+\delta t$, we can obtain the attributes of $\mathbf{p}$ at time $t$ by querying the space-time field with the point location at $t+\delta t$. 

In our framework, the motion network is conditioned on an embedding vector $\bm{\omega}$ (instead of queried timestep) and the motion can be written as $\Delta_{t\rightarrow t+\delta t}\mathbf{p}=M(\mathbf{p},\bm{\omega}_{t\rightarrow t+\delta t})$, where $\bm{\omega}_{t\rightarrow t+\delta t}$ depends on time $t$ and interval $\delta t$. Replacing the temporal variable $t$ with a vector $\bm{\omega}$ as input to $M$ enables predictability via embedding, as further detailed in \cref{sec:embed}.
All networks and the embedding vector \wrt time $t$ are optimized using the reconstruction loss $\mathcal{L}_{\mathrm{rec}}$ (\cf \cref{eq:rec}), with color $C^t$ predicted from $F,M,\bm{\omega}_{t\rightarrow t+\delta t}$ according to \cref{eq:rendering,eq:motion}.

We define the predictor $P$ as a function taking as input several motion embedding vectors of previous frames and inferring the motion embedding vectors for the future frames accordingly. Mathematically, we have 
$\bm{\omega}_{t\rightarrow t+\delta t}=P(\bm{\omega}_\mathrm{prev})$ with $\bm{\omega}_\mathrm{prev}=\{\bm{\omega}_{t-(i+1)\delta t \rightarrow t-i\delta t}\}_{i=1}^{\tau}$ set of $\tau$ previous frames' embeddings.
For example, in \cref{fig:overview}, the embedding vector $\bm{\omega}_{3\rightarrow 4}$ for motion from $t_3$ to $t_4$ is predicted from previous three embedding vectors, that is, $P\left(\{\bm{\omega}_{0\rightarrow 1},\bm{\omega}_{1\rightarrow 2},\bm{\omega}_{2\rightarrow 3}\}\right)$. 

\subsection{Neural Motion Fields with Motion Embedding}\label{sec:embed}
\begin{figure}[t]
    \centering
    \includegraphics[width=1.0\columnwidth,trim={0 0 0 0},clip]{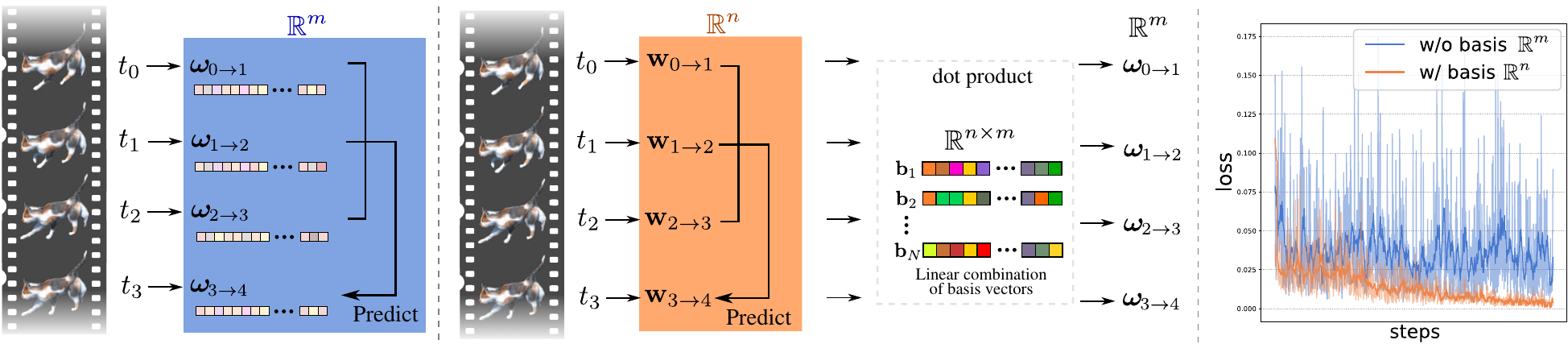}
     \vspace{-16pt}
    \caption{We use a set of basis vectors for the motion embedding (middle), rather than associating each frame with a motion vector (left). The input and out space of the predictor switches to the linear combination weights by using these shared basis vectors. The comparison of training losses (right) indicates that the predictor converges faster on the space of linear combination weights.
    }
    \label{fig:basis}
\end{figure}

The motion field is conditioned on an embedding vector, sampled from a latent space depicting motion patterns. Such embedding can be implemented in various ways. The simplest one is to associate each motion of interest with a trainable embedding vector. This technique has been widely used for conditioning neural fields \wrt appearance \cite{martin2021nerf} and deformation \cite{Park_2021_ICCV}. 
However, empirical studies show that associating each motion with motion embedding frequently and significantly slows down the convergence speed of the predictor, as demonstrated in \cref{fig:basis}. 
We presume that the phenomenon is caused by the large and unstructured solution space brought by frame-wise motion embedding. To validate the assumption and improve the convergence speed, we propose to reduce the dimension of the input and output space of the predictor.

Inspired by mixture-of-experts-based prediction networks \cite{starke2019neural,ling2020character,hassan2021stochastic}, 
we design a set $\mathbf{B}\in\mathbb{R}^{n\times m}$ of $n$ embedding basis vectors, \ie, $\mathbf{B} = [\mathbf{b}_1,\cdots,\mathbf{b}_n]^T$ with $\mathbf{b}_i\in\mathbb{R}^m$ basis vector. 
$\mathbf{B}$ is shared across all frames. 
Then the motion embedding becomes 
$\bm{\omega}_{t \rightarrow t + \delta t}= \mathbf{w}_{t \rightarrow t + \delta t} \cdot \mathbf{B}$, with $\mathbf{w} \in \mathbb{R}^n$
optimizable linear combination weights.
Accordingly, we redefine the model $P$ to receive and predict these weight vectors instead of the embedding ones, thus reducing its
input space and output space to $\mathbb{R}^n$, \ie, with the dimensionality of basis vectors not affecting the predictor anymore. In our experiments, we set $n=5$ and $m=32$,
so the dimension of the predictor's output space is reduced from 32 to 5. 
An illustration and comparison of the training losses between the two schemes are presented in \cref{fig:basis}.

\subsection{Regularizing with Motion Prediction}\label{sec:pred}
Our proposed solution makes it possible to complement the usual self-supervision of space-time neural fields (through visual reconstruction only) by a regularization term over motion.
However, while predicting motion embeddings is straightforward, \ie, by simply forwarding the embedding vectors of previous frames into $P$, leveraging $P$ for the regularization of $M$ is not trivial. 

In our framework, motion embeddings can be acquired either from reconstruction, \ie, optimizing each embedding along with other components (\eg, both the motion embedding $\bm{\omega}_{3\rightarrow 4} = \mathbf{w}_{3\rightarrow 4}\cdot \mathbf{B}$ and the motion network $M(\mathbf{p},\bm{\omega}_{3\rightarrow 4})$ can be optimized on observed images at $t=3,4$); or through the predictor (\eg, $\bm{\omega}_{3\rightarrow 4} = P(\left\{\mathbf{w}_{t-1 \rightarrow t}\right\}_{t=1}^{3})\cdot \mathbf{B}$).
%
We leverage this redundancy for regularization, \ie, proposing a loss to minimize the difference between the self-supervised embeddings and their corresponding predicted versions:
\begin{equation}
\mathcal{L}_{\mathrm{pred}}=\| 
P\left(\mathbf{w}_{\mathrm{prev}}\right) 
- 
\argmin_{\mathbf{w}_{t \rightarrow t + \delta t}} \mathcal{L}_{\mathrm{rec}}
\|_2^2%
, \text{~where~} 
\mathbf{w}_\mathrm{prev}=\{\mathbf{w}_{t-(i+1)\delta t \rightarrow t-i\delta t}\}_{i=1}^{\tau}.
\label{eq:l_pred}
\end{equation}
In the above equation, the first term $P(\cdot)$ represents the motion embedding predicted according to previous $\tau$ frames, 
and the second term $\argmin_{\mathbf{w}_{t \rightarrow t + \delta t}} \mathcal{L}_{\mathrm{rec}}$ is the vector acquired from minimizing the reconstruction loss. 

It is, however, impractical to compute this second term during training, since the reconstruction problem can take hours to solve via optimization.
We propose instead to obtain $\mathbf{w}_{{t \rightarrow t + \delta t}}$ in an online manner, and to jointly optimize frame weights $\mathbf{w}$ over both $\mathcal{L}_{\mathrm{rec}}$ and $\mathcal{L}_{\mathrm{pred}}$ at each optimization step.
That is, at each step, all current frame weights $\mathbf{w}$ are first used to compute $\mathcal{L}_{\mathrm{pred}}$ and optimize downstream models accordingly, and are then themselves optimized \wrt $\mathcal{L}_{\mathrm{rec}}$.
The details of implementing the two losses with batches of frames are introduced in the next section.

\subsection{Optimization}

\begin{figure}[t]
    \centering
    \includegraphics[width=0.8\columnwidth,trim={0 0 0 0},clip]{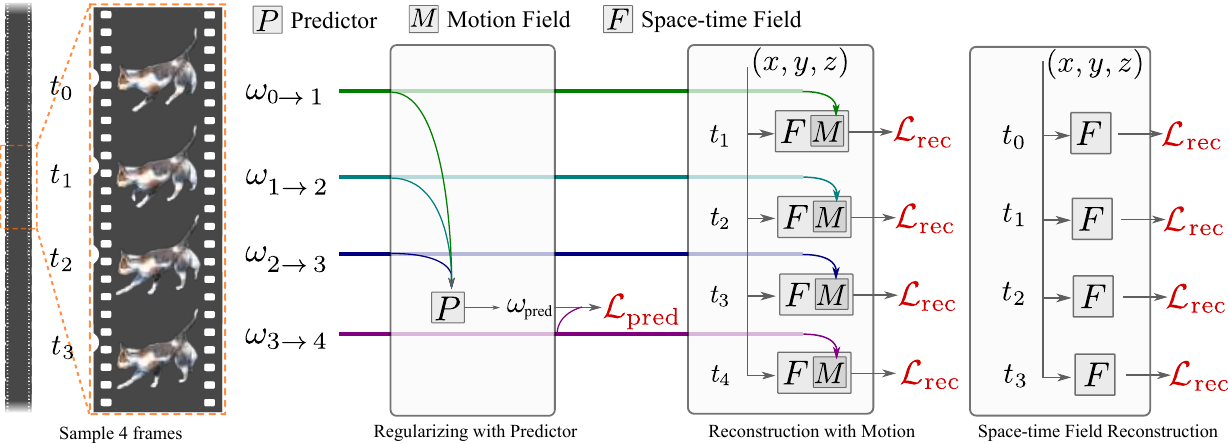}
    \caption{System optimization, demonstrated on a batch of 4 frames. Predictor $P$ infers a vector $\bm{\omega}$ based on the preceding 3 frames; $\mathcal{L}_{\mathrm{pred}}$ minimizes the difference between these predicted embeddings and their sampled equivalents; whereas reconstruction loss $\mathcal{L}_{\mathrm{rec}}$ is applied to the predicted four frames, with and without motion reparameterization. 
    }
    \label{fig:optim}
\end{figure}

During optimization, we sample a short sequence of frames from the training set. For simplifying the notations, we assume that the predictor takes $\tau=3$ frames as input and predicts the motion of the next frame. 
An illustration is presented in \cref{fig:optim}. Four consecutive frames $(t_i,t_{i+1},t_{i+2},t_{i+3})$ are first sampled from the observed sequence and the corresponding embedding vectors $\bm{\omega}$ are acquired as in \cref{sec:embed}. 
Note that training images can be sampled from different synchronized cameras if available.

We disentangle appearance- and motion-related information during optimization by applying $\mathcal{L}_{\mathrm{rec}}$ to images reconstructed both with and without motion reparameterization. That is, we sample $F$ for radiance/density values $(\mathbf{c}_t, \bm{\sigma}_t)$ both as $F\left(\mathbf{p} + M(\mathbf{p}, \bm{\omega}_{t \rightarrow t + \delta t}), t + \delta t\right)$
and as
$F\left(\mathbf{p}, t\right)$ 
(\cf, \cref{fig:optim}).

\section{Experiments}
We qualitatively and quantitatively evaluate our method in this section. 
\emph{We urge the reader to check our video to better appraise the quality of motion.} 
The following three datasets are used for evaluation:
\begin{itemize}
    \item \textbf{ZJU-MoCap} \cite{peng2021neural} is a multi-camera dataset, with videos of one person performing different actions. Since each video sequence records a single human, the scene is less topologically varying and we compare our method with canonical frame-based representations of dynamic scenes. We use videos from 11 cameras for evaluation.
    \item \textbf{Panoptic} \cite{joo2015panoptic} includes videos from multiple synchronized cameras under many different settings including multi-person activities and human-object interactions. We select 4 challenging and representative video clips from the 31 HD cameras and denote them as \textsc{Sports, Tools, Ian}, and \textsc{Cello}. Each clip has 400 frames and all the clips involve human-object interaction.
    \item \textbf{Hypernerf} \cite{park2021hypernerf} is a single-camera dataset, \ie, with one view available at each timestamp. Unlike the previous two datasets that use static cameras, in Hypernerf the multiview information is generated by moving the camera around. Hypernerf is challenging not only because of the single-camera setting, but also the topologically varying scenes.
\end{itemize}
Details about the clips (\eg, starting and ending frame number) are included in the supplementary. All the sequences are split into short intervals consisting of 25 frames. On each interval, the networks are trained using an Adam optimizer \cite{KingmaB14} with a learning rate that decays from $5\times10^{-4}$ to $5\times10^{-6}$ every 50k iterations. 
During training, the two losses are added with a balancing parameter, \ie, $\mathcal{L}=\mathcal{L}_{\mathrm{rec}}+\gamma \mathcal{L}_{\mathrm{pred}}$  with $\gamma$ set to 0.01 in all experiments. A batch of 1,024 rays is randomly sampled from the selected frames for training the motion field and the space-time field. We observe that using viewing direction $\mathbf{d}$ in $F$ leads to worse performance if the scene of interest mostly contains Lambertian surfaces. In our experiments, the viewing direction is not taken as the input for the space-time field, \ie, a space-time irradiance field \cite{xian2020space}. 
The network structures of the motion field and the space-time field are the same as in NeRF \cite{mildenhall2020nerf}. The predictor consists of 5 fully connected layers with a width of 128 and ReLU activations.


\begin{figure}[tb]
    \centering
    \includegraphics[width=0.9\columnwidth,trim={0 0 0 0},clip]{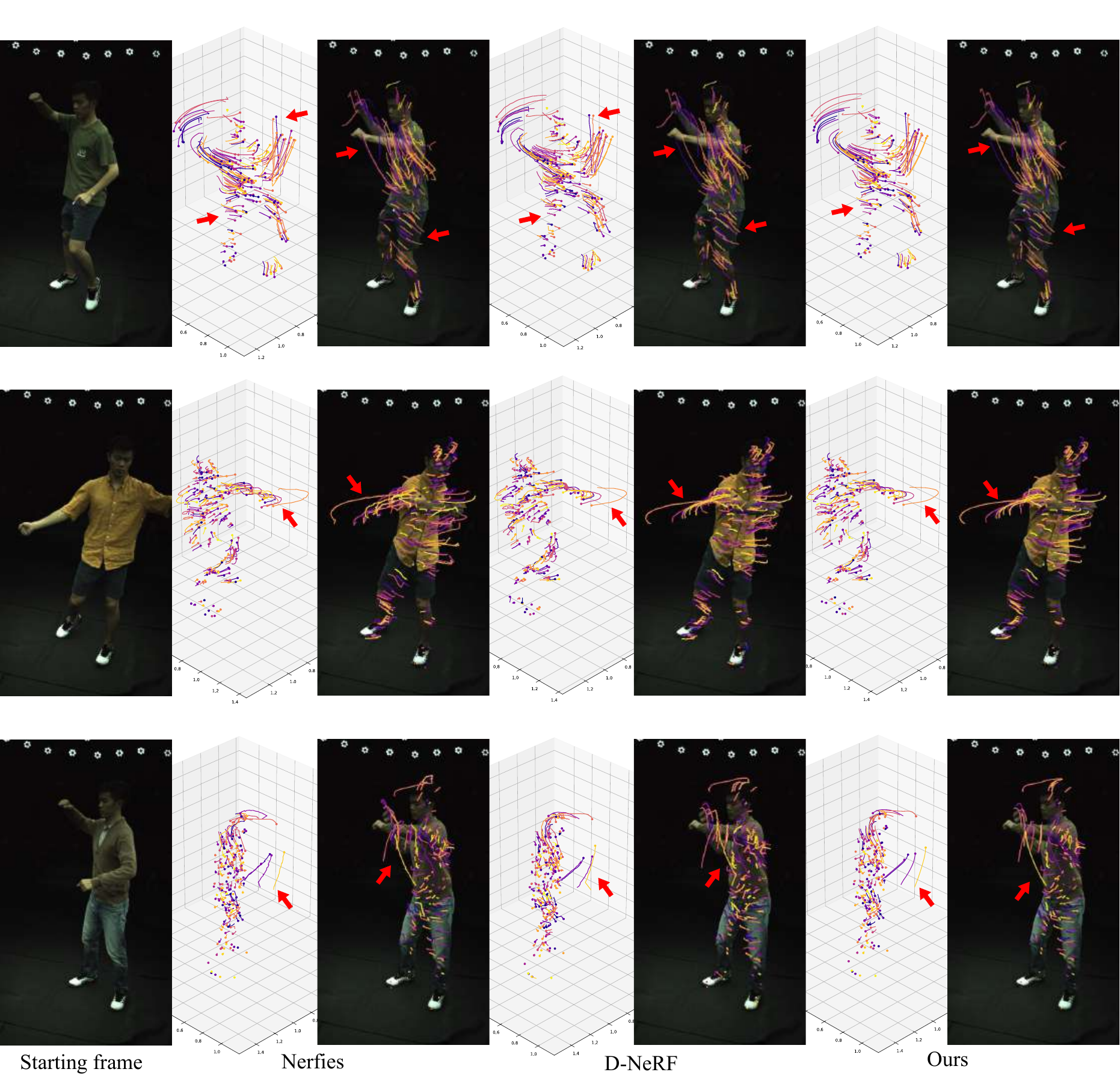}
    \vspace{-16pt}
    \caption{Comparison of the estimated motion on the ZJU-MoCap dataset. Only one person is captured for each sequence and we compare our method with canonical frame-based methods Nerfies \cite{Park_2021_ICCV} and D-NeRF \cite{pumarola2020d}. Motion for 20 frames is demonstrated.}
    \label{fig:zju}
\end{figure}

\begin{figure}[p]
    \centering
    \includegraphics[width=1.0\columnwidth,trim={0 0 0 25},clip]{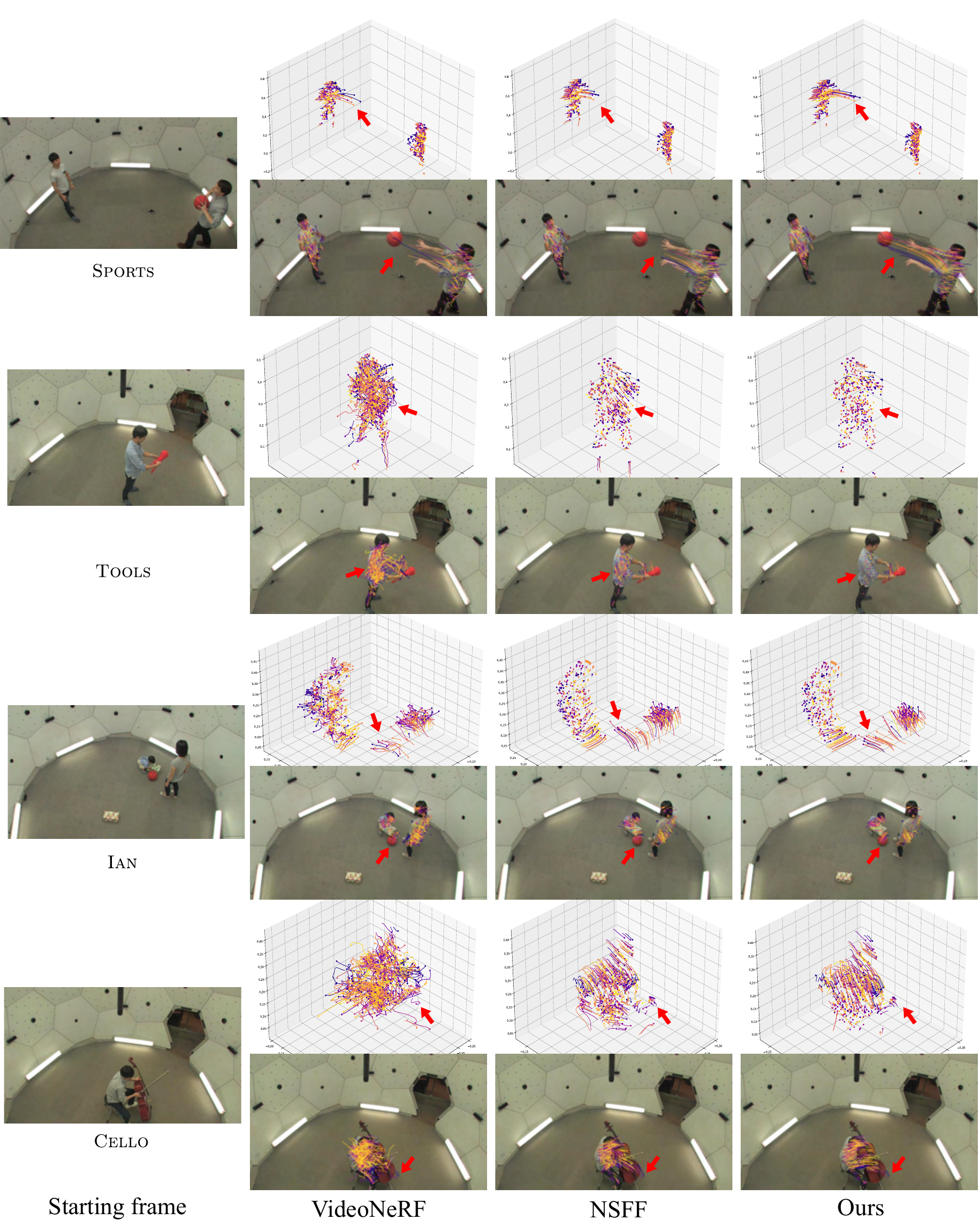}
    \caption{Motion estimation comparison on the Panoptic dataset  \cite{joo2015panoptic}.
    Motions estimated by VideoNeRF are more chaotic than NSFF, possibly due to the 2D optical flow supervision adopted in NSFF.
    Our method faithfully estimates the motions of people and objects, whereas NSFF fails to track some points, \eg, the ball in \textsc{Sports} and \textsc{Ian}.
    }
    \label{fig:panop}
\end{figure}

\begin{figure}[tb]
    \centering
    \includegraphics[width=1.0\columnwidth,trim={0 0 0 0},clip]{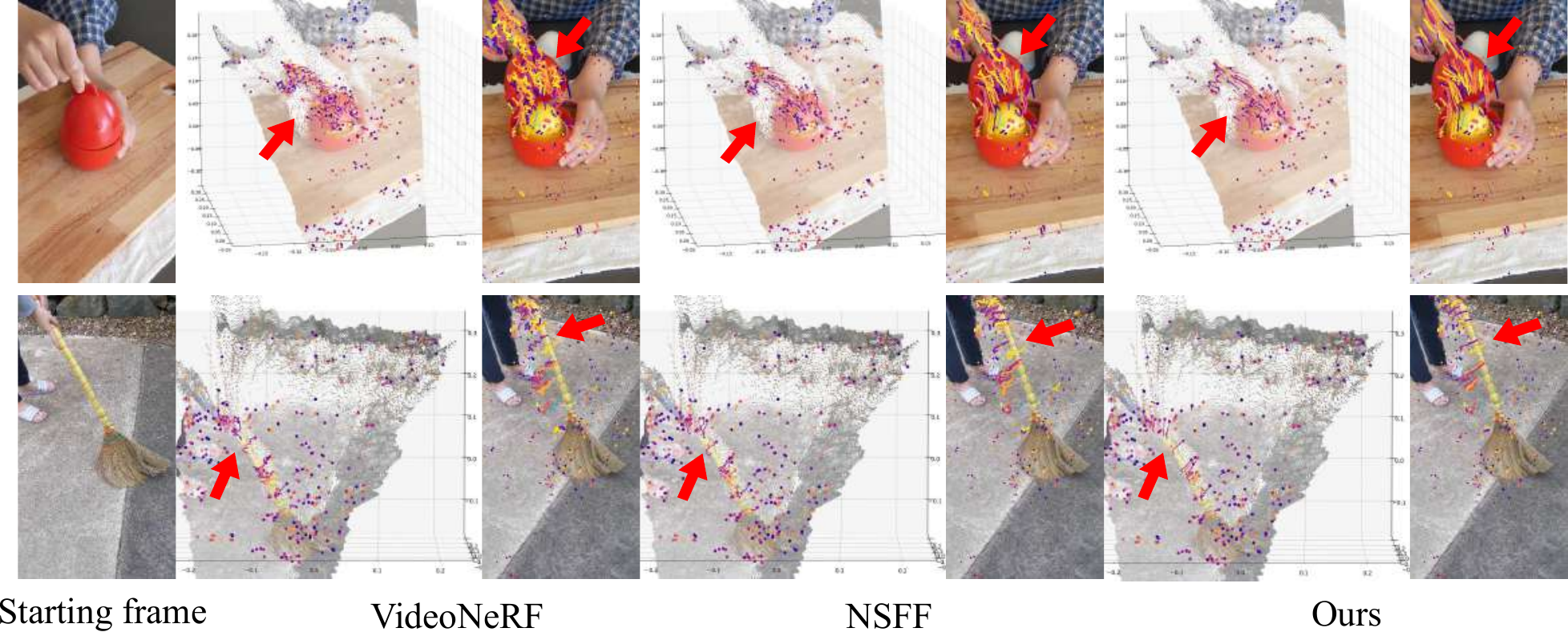}
    \caption{Comparison of the estimated motion on the Hypernerf dataset \cite{park2021hypernerf}.
    We randomly sample points on the surfaces and then demonstrate their the motions.
    }
    \label{fig:hypernerf}
\end{figure}


\subsection{Qualitative Evaluation}
We visually compare the estimated motion in this section. 
Since neural motion fields tracks all points in the space, we randomly sample points and then demonstrate their trajectory. Different sampling strategies are used for different datasets. For ZJU-MoCap, we first sample a dense grid of points and then remove the empty points with $\sigma<20$, then we randomly sample points from the non-empty ones. 
For Panoptic, since background  (walls and floors) is kept in the scenes, we sample meaningful points near the persons in the scene, leveraging provided people positions. 
For Hypernerf, since the scenes are all front-facing, we sample points on the surfaces according to the depth generated by the space-time field $F$ from one view. 

\subsubsection{On Multi-camera Dataset.}
We first present our results on ZJU-MoCap in \cref{fig:zju}. Since there is only one person in this dataset, the topology of the scene roughly remains unchanged and canonical space-based methods can be applied. Nerfies \cite{Park_2021_ICCV} and D-NeRF \cite{pumarola2020d} are selected for comparison. As can be observed from the images, our method can generate a  smooth motion as opposed to the rugged and noisy motions from the other two methods.

\cref{fig:panop} demonstrate the performance of our method and competitors on the Panoptic dataset. The scenes contain complex geometries and objects may occur or disappear in the middle of a sequence. Two space-time field-based methods, VideoNeRF \cite{xian2020space} and NSFF \cite{li2020neural}, are selected for comparison. Our method estimate the motion of both people and objects accurately, while VideoNeRF presents chaotic results and the motion from NSFF are occasionally inaccurate. 
The results on \cref{fig:zju,fig:panop} validate our claim that our method can well track all points in the space without prior knowledge of the scene.

\subsubsection{On Single-camera Dataset.}
To further validate our method, we demonstrate motion estimation in single-camera settings, which are more commonly encountered by dynamic-scene novel-view rendering methods. We consider the challenging scenes captured by Hypernerf \cite{park2021hypernerf}. As shown in \cref{fig:hypernerf}, we compare again to VideoNeRF \cite{xian2020space} and NSFF \cite{li2020neural}.
We note that our results are more temporally consistent and accurate than competitors. These results highlight the practical value of our method, able to accurately handle single-camera image sequences captured in the wild.

\subsection{Quantitative Evaluation}
Quantitative evaluation is difficult for our task since manually labeling a dense set of points in the space is expensive, if not unfeasible. 
We thus use the sparser human body joints provided by the Panoptic dataset to quantify the accuracy of the estimated motion.
MPJPE \cite{sigal2010humaneva} and 3D-PCK \cite{mehta2017monocular} are two widely used metrics for evaluating 3D human pose tracking performance, but both of them do not suit our task since our tracking requires as input the position of points at the starting frame.
We propose to calculate the tracking error across $K$ frames and use the averaged value as a metric. We denote the metric as $\mathrm{mMPJPE}_K$ (mean MPJPE), computed as:
\begin{equation}
\mathrm{mMPJPE}_K = \frac{1}{N_f} \frac{1}{K} \sum_{u=1}^{N_f} \sum_{v=u+1}^{u+K} \mathrm{MPJPE}(P_{u\rightarrow v},P_v^{\mathrm{gt}}),
\end{equation}
where $K$ is the number of frames for evaluating the motion and $N_f$ is the total number of frames in the sequence. $P_{i\rightarrow j}$ represents the estimated positions for the $j$th frame given positions for the $i$th one as inputs, and $P_j^{\mathrm{gt}}$ the ground-truth joint positions for the $j$th frame.

\begin{figure}[tb]
    \centering
    \includegraphics[width=0.8\columnwidth,trim={0 0 0 26},clip]{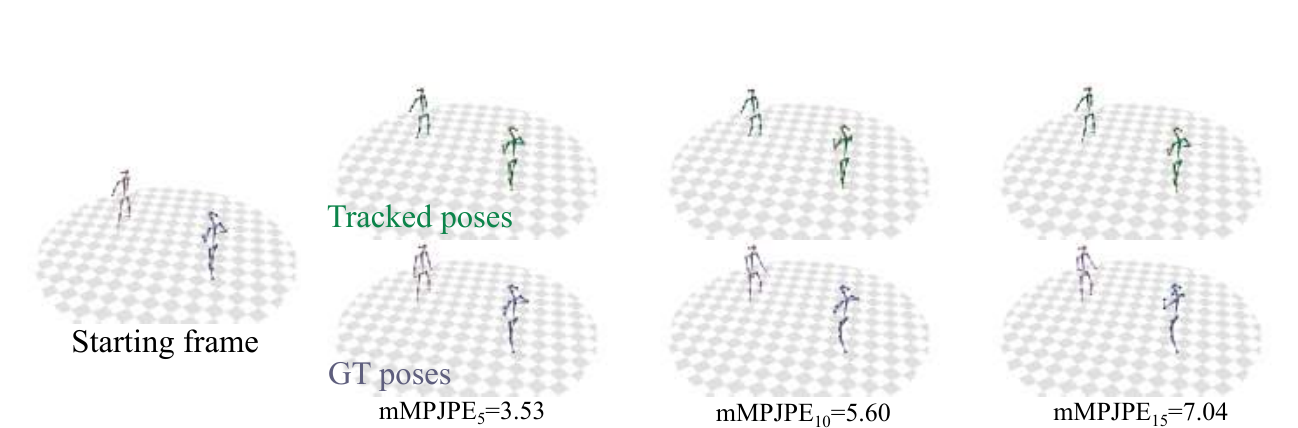}
    \caption{Visualization of motion tracking results on the Panoptic dataset.}
    \label{fig:mpjpe}
\end{figure}

We report the mMPJPEK metric with $K =$ 5, 10, 15 on the Panoptic dataset
in \cref{tab:tracking}. Our method achieves more accurate tracking performance than the other two methods except on \textsc{Ian} while tracking with 5 and 10 frames. NSFF requires both 2D optical flow and depth, while VideoNeRF requires depth information. 
As a comparison, we do not use any data-driven prior to guide the motion estimation module.
Moreover, in \cref{fig:mpjpe} we visualize the tracked pose and the ground truth pose on one sequence and compute the corresponding mMPJPE metrics ($N_f=1$ for one sequence).

\begin{table}[t]
    \footnotesize
    \centering
    \caption{Quantitatively evaluating the estimated motion on the Panoptic dataset. Locations of the body joints in the starting frame are used as the inputs and we calculate the averaged tracking error for the body joints.
    }
    \resizebox{\textwidth}{!}{
    \begin{tabular}{c|ccc|ccc|ccc}
    \toprule
    & \multicolumn{3}{c|}{$\mathrm{mMPJPE}_{5}$ (cm)} 
    & \multicolumn{3}{c|}{$\mathrm{mMPJPE}_{10}$ (cm)} 
    & \multicolumn{3}{c}{$\mathrm{mMPJPE}_{15}$ (cm)} \\
& ~\scriptsize{VideoNeRF} & \scriptsize{~~~NSFF~~} & \scriptsize{~~~Ours~~} 
& ~\scriptsize{VideoNeRF} & \scriptsize{~~~NSFF~~} & \scriptsize{~~~Ours~~} 
& ~\scriptsize{VideoNeRF} & \scriptsize{~~~NSFF~~} & \scriptsize{~~~Ours~~} 
    \\ \midrule
    \textsc{Sports}~~ & 5.942 & 5.171 & \textbf{4.533} & 8.346 & 7.933 & \textbf{7.457} & 11.569 & 11.254 & \textbf{10.718} \\
    \textsc{Tools}~~ & 3.378 & 2.341 & \textbf{1.684} & 4.105 & 2.879 & \textbf{2.650} & 4.931 & 3.984 & \textbf{3.393} \\
    \textsc{Ian}~~ & 3.448 & \textbf{2.349} & 2.402 & 5.059 & \textbf{3.534} & 3.792 & 6.767 & 5.282 & \textbf{4.980} \\
    \textsc{Cello}~~ & 2.796 & 1.759 & \textbf{1.612} & 4.281 & 3.296 & \textbf{2.572} & 4.853 & 3.776 & \textbf{3.457} \\
    \bottomrule
    \end{tabular}
    }
    \label{tab:tracking}
\end{table}

\begin{figure}[tb]
    \centering
    \includegraphics[width=1.0\columnwidth,trim={0 0 0 0},clip]{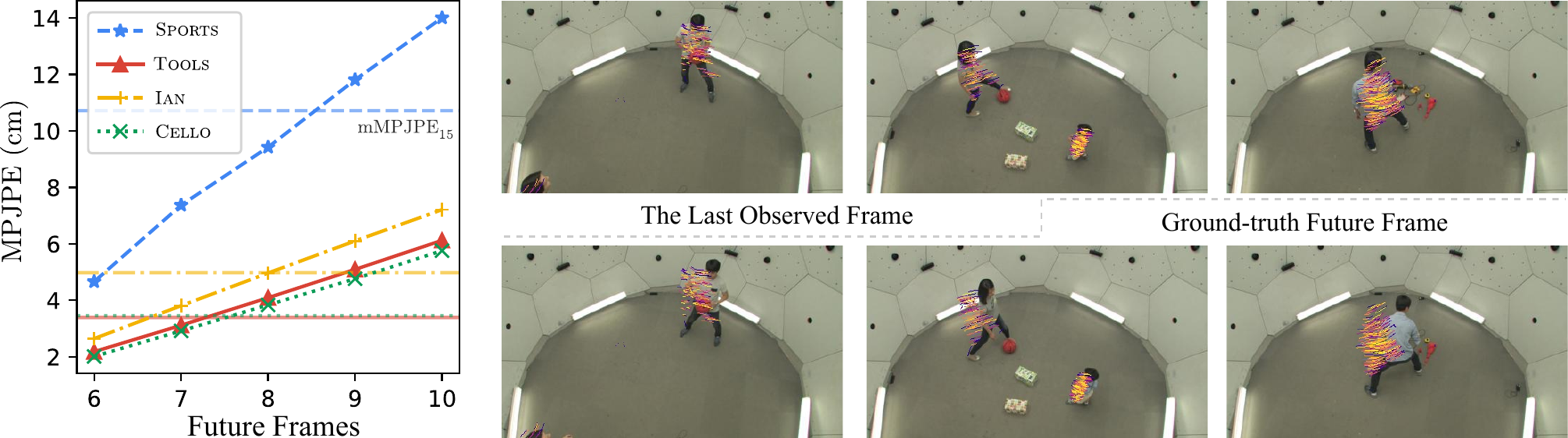}
    \caption{Accuracy evaluation of the motion predictor. Left: Plotting of the MPJPE of predicted future body joint locations. Horizontal lines are the mMPJPE$_{15}$ results on the corresponding scenes. Right: Visualization of predicted future motion of densely sampled points on the last observed frames and the \nth{10} ground-truth future frames.}
    \label{fig:predictor}
\end{figure}

\subsection{Analysis of the Motion Predictor}
We analyze the motion predictor $P$ in two aspects: prediction accuracy and transferability. 
For the accuracy evaluation, we compare the predicted future locations of the body joints and the ground-truth future locations. 
The results are demonstrated in \cref{fig:predictor}. The training sequences are separated into 20 intervals and we test the prediction results on each interval. The MPJPE of predicted body joint locations are averaged over all the intervals and plotted. 
We can observe in \cref{tab:tracking} that the model can predict the unseen motion of the next 5 time steps, with a low error close to the the tracking error over actual observations.

\begin{figure}[tb]
    \centering
    \includegraphics[width=0.8\columnwidth,trim={0 0 0 0},clip]{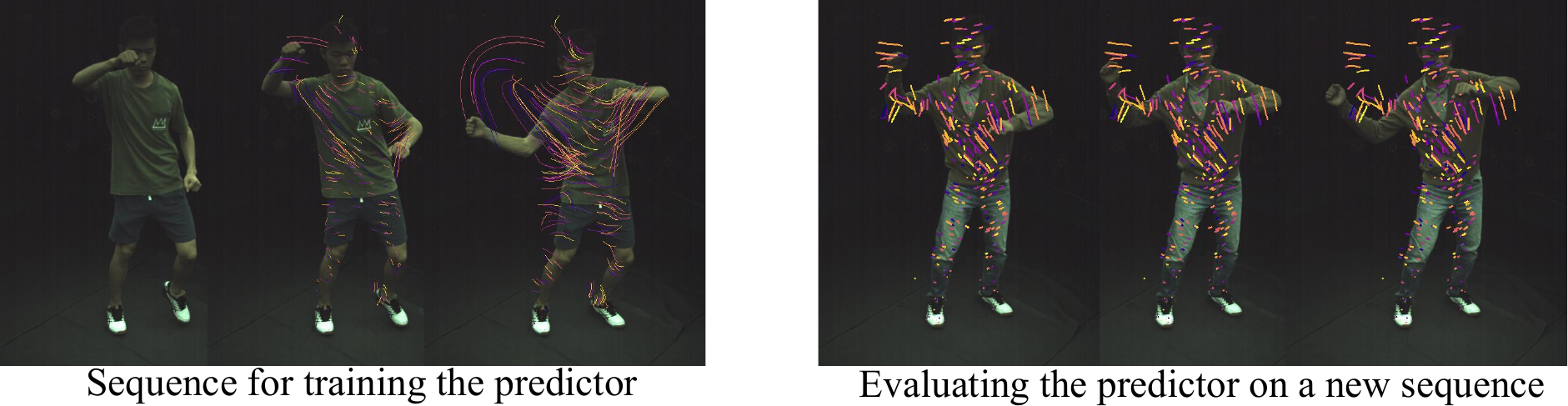}
    \caption{Transferability of the motion predictor. We train the whole framework on the left-side sequence, then we freeze the predictor and fine tune other models on the right-side sequence. The next 10 frame motions are predicted from the last observed frame (the right-side first image) and visualized. The other two images are real movements in the future \nth{5} frame and \nth{10} frame.
    }
    \label{fig:pred-trans}
\end{figure}

We further demonstrate the transferability of the predictor in \cref{fig:pred-trans}. Since the predictor generates 
motion codes in a latent space, the same model should work for motion sequences with similar patterns. We test the intuition on the ZJU-MoCap dataset, on two sequences in which the person does similar actions. We can observe from the right side of the figure that the predicted motions align with the real movements. The results demonstrate that the predictor is indeed transferable if the motions are similar.


\section{Discussion}
\paragraph{\textbf{Limitations.}} Our method sometimes fail on non-rigid/monochromatic elements and the problem of motion estimation then gets underconstrained: Some points may converge into the same point for the non-rigid case and it may be hard to tell which part in the monochromatic area moved. We presume that a more advanced (possibly pre-trained) motion prediction model could be leveraged. Moreover, while our methods shows higher precision in estimating natural motion (\eg, dense human motion tracking), it is among our future work to address some other challenging scenes (\eg, scenes with chaotic particles). 

\paragraph{\textbf{Conclusion.}} We introduced a novel solution for the regularization and prediction of 3D dense motion in dynamic scenes. Leveraging advances in neural fields, we propose a combination of space-time and motion fields conditioned on motion embeddings. Through predictability-based regularization over these embeddings, we promote the encoding of scene-relevant motions and penalize ambiguous and noisy deformations. 
We acknowledge that this scheme may not benefit all types of scenes (\cf above limitations), but it shows higher precision in natural settings.


\clearpage
%
%
\bibliographystyle{splncs04}
\bibliography{egbib}

\begin{thebibliography}{10}
\providecommand{\url}[1]{\texttt{#1}}
\providecommand{\urlprefix}{URL }
\providecommand{\doi}[1]{https://doi.org/#1}

\bibitem{basha2013multi}
Basha, T., Moses, Y., Kiryati, N.: Multi-view scene flow estimation: A view
  centered variational approach. International journal of computer vision
  \textbf{101}(1),  6--21 (2013)

\bibitem{boss2021nerd}
Boss, M., Braun, R., Jampani, V., Barron, J.T., Liu, C., Lensch, H.: Nerd:
  Neural reflectance decomposition from image collections. In: Proceedings of
  the IEEE/CVF International Conference on Computer Vision. pp. 12684--12694
  (2021)

\bibitem{boss2021neuralpil}
Boss, M., Jampani, V., Braun, R., Liu, C., Barron, J.T., Lensch, H.P.:
  Neural-pil: Neural pre-integrated lighting for reflectance decomposition. In:
  Advances in Neural Information Processing Systems (2021)

\bibitem{box1977canonical}
Box, G.E., Tiao, G.C.: A canonical analysis of multiple time series. Biometrika
   \textbf{64}(2),  355--365 (1977)

\bibitem{bozic2020neural}
Bozic, A., Palafox, P., Zollh{\"o}fer, M., Dai, A., Thies, J., Nie{\ss}ner, M.:
  Neural non-rigid tracking. Advances in Neural Information Processing Systems
  \textbf{33},  18727--18737 (2020)

\bibitem{tensorf}
Chen, A., Xu, Z., Geiger, A., Yu, J., Su, H.: Tensorf: Tensorial radiance
  fields. In: Proceedings of the European Conference on Computer Vision (2022)

\bibitem{chibane2020neural}
Chibane, J., Pons-Moll, G., et~al.: Neural unsigned distance fields for
  implicit function learning. Advances in Neural Information Processing Systems
   \textbf{33} (2020)

\bibitem{chung2018survey}
Chung, S.J., Paranjape, A.A., Dames, P., Shen, S., Kumar, V.: A survey on
  aerial swarm robotics. IEEE Transactions on Robotics  \textbf{34}(4),
  837--855 (2018)

\bibitem{du2021nerflow}
Du, Y., Zhang, Y., Yu, H.X., Tenenbaum, J.B., Wu, J.: Neural radiance flow for
  4d view synthesis and video processing. In: Proceedings of the IEEE/CVF
  International Conference on Computer Vision (2021)

\bibitem{fang2022fast}
Fang, J., Yi, T., Wang, X., Xie, L., Zhang, X., Liu, W., Nie{\ss}ner, M., Tian,
  Q.: Fast dynamic radiance fields with time-aware neural voxels. arXiv
  preprint arXiv:2205.15285  (2022)

\bibitem{gafni2020dynamic}
Gafni, G., Thies, J., Zollhofer, M., Nie{\ss}ner, M.: Dynamic neural radiance
  fields for monocular 4d facial avatar reconstruction. In: Proceedings of the
  IEEE/CVF Conference on Computer Vision and Pattern Recognition. pp.
  8649--8658 (2021)

\bibitem{goerg2013forecastable}
Goerg, G.: Forecastable component analysis. In: International conference on
  machine learning. pp. 64--72. PMLR (2013)

\bibitem{hassan2021stochastic}
Hassan, M., Ceylan, D., Villegas, R., Saito, J., Yang, J., Zhou, Y., Black,
  M.J.: Stochastic scene-aware motion prediction. In: Proceedings of the
  IEEE/CVF International Conference on Computer Vision. pp. 11374--11384 (2021)

\bibitem{hong2021headnerf}
Hong, Y., Peng, B., Xiao, H., Liu, L., Zhang, J.: Headnerf: A real-time
  nerf-based parametric head model. In: Proceedings of the IEEE/CVF Conference
  on Computer Vision and Pattern Recognition. pp. 20374--20384 (2022)

\bibitem{Huang2022PREF}
Huang, B., Yan, X., Chen, A., Gao, S., Yu, J.: Pref: Phasorial embedding fields
  for compact neural representations  (2022)

\bibitem{huang1981image}
Huang, T.S., Tsai, R.: Image sequence analysis: Motion estimation. In: Image
  sequence analysis, pp. 1--18. Springer (1981)

\bibitem{jiang2020shapeflow}
Jiang, C., Huang, J., Tagliasacchi, A., Guibas, L.: Shapeflow: Learnable
  deformations among 3d shapes. In: Advances in Neural Information Processing
  Systems (2020)

\bibitem{joo2015panoptic}
Joo, H., Liu, H., Tan, L., Gui, L., Nabbe, B., Matthews, I., Kanade, T.,
  Nobuhara, S., Sheikh, Y.: Panoptic studio: A massively multiview system for
  social motion capture. In: Proceedings of the IEEE International Conference
  on Computer Vision. pp. 3334--3342 (2015)

\bibitem{KingmaB14}
Kingma, D.P., Ba, J.: Adam: {A} method for stochastic optimization. In: Bengio,
  Y., LeCun, Y. (eds.) International Conference on Learning Representations
  (2015)

\bibitem{li2021neural}
Li, T., Slavcheva, M., Zollhoefer, M., Green, S., Lassner, C., Kim, C.,
  Schmidt, T., Lovegrove, S., Goesele, M., Newcombe, R., et~al.: Neural 3d
  video synthesis from multi-view video. In: Proceedings of the IEEE/CVF
  Conference on Computer Vision and Pattern Recognition. pp. 5521--5531 (2022)

\bibitem{li2020neural}
Li, Z., Niklaus, S., Snavely, N., Wang, O.: Neural scene flow fields for
  space-time view synthesis of dynamic scenes. In: Proceedings of the IEEE/CVF
  Conference on Computer Vision and Pattern Recognition (2021)

\bibitem{li2017robust}
Li, Z., Ji, Y., Yang, W., Ye, J., Yu, J.: Robust 3d human motion reconstruction
  via dynamic template construction. In: International Conference on 3D Vision.
  pp. 496--505. IEEE (2017)

\bibitem{ling2020character}
Ling, H.Y., Zinno, F., Cheng, G., Van De~Panne, M.: Character controllers using
  motion vaes. ACM Transactions on Graphics (TOG)  \textbf{39}(4),  40--1
  (2020)

\bibitem{liu2021neural}
Liu, L., Habermann, M., Rudnev, V., Sarkar, K., Gu, J., Theobalt, C.: Neural
  actor: Neural free-view synthesis of human actors with pose control. ACM
  Trans. Graph.(ACM SIGGRAPH Asia)  (2021)

\bibitem{lombardi2019neural}
Lombardi, S., Simon, T., Saragih, J., Schwartz, G., Lehrmann, A., Sheikh, Y.:
  Neural volumes: learning dynamic renderable volumes from images. ACM
  Transactions on Graphics  \textbf{38}(4),  1--14 (2019)

\bibitem{martin2021nerf}
Martin-Brualla, R., Radwan, N., Sajjadi, M.S., Barron, J.T., Dosovitskiy, A.,
  Duckworth, D.: Nerf in the wild: Neural radiance fields for unconstrained
  photo collections. In: Proceedings of the IEEE/CVF Conference on Computer
  Vision and Pattern Recognition. pp. 7210--7219 (2021)

\bibitem{mehta2017monocular}
Mehta, D., Rhodin, H., Casas, D., Fua, P., Sotnychenko, O., Xu, W., Theobalt,
  C.: Monocular 3d human pose estimation in the wild using improved cnn
  supervision. In: 2017 international conference on 3D vision (3DV). pp.
  506--516. IEEE (2017)

\bibitem{menze2015object}
Menze, M., Geiger, A.: Object scene flow for autonomous vehicles. In:
  Proceedings of the IEEE conference on computer vision and pattern
  recognition. pp. 3061--3070 (2015)

\bibitem{mescheder2019occupancy}
Mescheder, L., Oechsle, M., Niemeyer, M., Nowozin, S., Geiger, A.: Occupancy
  networks: Learning 3d reconstruction in function space. In: Proceedings of
  the IEEE/CVF Conference on Computer Vision and Pattern Recognition. pp.
  4460--4470 (2019)

\bibitem{mildenhall2020nerf}
Mildenhall, B., Srinivasan, P.P., Tancik, M., Barron, J.T., Ramamoorthi, R.,
  Ng, R.: Nerf: Representing scenes as neural radiance fields for view
  synthesis. In: European Conference on Computer Vision. pp. 405--421. Springer
  (2020)

\bibitem{mittal2020just}
Mittal, H., Okorn, B., Held, D.: Just go with the flow: Self-supervised scene
  flow estimation. In: Proceedings of the IEEE/CVF conference on computer
  vision and pattern recognition. pp. 11177--11185 (2020)

\bibitem{newcombe2015dynamicfusion}
Newcombe, R.A., Fox, D., Seitz, S.M.: Dynamicfusion: Reconstruction and
  tracking of non-rigid scenes in real-time. In: Proceedings of the IEEE
  conference on computer vision and pattern recognition. pp. 343--352 (2015)

\bibitem{niemeyer2019occupancy}
Niemeyer, M., Mescheder, L., Oechsle, M., Geiger, A.: Occupancy flow: 4d
  reconstruction by learning particle dynamics. In: Proceedings of the IEEE/CVF
  international conference on computer vision. pp. 5379--5389 (2019)

\bibitem{Noguchi_2021_ICCV}
Noguchi, A., Sun, X., Lin, S., Harada, T.: Neural articulated radiance field.
  In: Proceedings of the IEEE/CVF International Conference on Computer Vision.
  pp. 5762--5772 (October 2021)

\bibitem{park2019deepsdf}
Park, J.J., Florence, P., Straub, J., Newcombe, R., Lovegrove, S.: Deepsdf:
  Learning continuous signed distance functions for shape representation. In:
  Proceedings of the IEEE/CVF Conference on Computer Vision and Pattern
  Recognition. pp. 165--174 (2019)

\bibitem{Park_2021_ICCV}
Park, K., Sinha, U., Barron, J.T., Bouaziz, S., Goldman, D.B., Seitz, S.M.,
  Martin-Brualla, R.: Nerfies: Deformable neural radiance fields. In:
  Proceedings of the IEEE/CVF International Conference on Computer Vision. pp.
  5865--5874 (October 2021)

\bibitem{park2021hypernerf}
Park, K., Sinha, U., Hedman, P., Barron, J.T., Bouaziz, S., Goldman, D.B.,
  Martin-Brualla, R., Seitz, S.M.: Hypernerf: A higher-dimensional
  representation for topologically varying neural radiance fields. ACM Trans.
  Graph.  \textbf{40}(6) (dec 2021)

\bibitem{pena1987identifying}
Pena, D., Box, G.E.: Identifying a simplifying structure in time series.
  Journal of the American statistical Association  \textbf{82}(399),  836--843
  (1987)

\bibitem{Peng_2021_ICCV}
Peng, S., Dong, J., Wang, Q., Zhang, S., Shuai, Q., Zhou, X., Bao, H.:
  Animatable neural radiance fields for modeling dynamic human bodies. In:
  Proceedings of the IEEE/CVF International Conference on Computer Vision. pp.
  14314--14323 (October 2021)

\bibitem{peng2021neural}
Peng, S., Zhang, Y., Xu, Y., Wang, Q., Shuai, Q., Bao, H., Zhou, X.: Neural
  body: Implicit neural representations with structured latent codes for novel
  view synthesis of dynamic humans. In: Proceedings of the IEEE/CVF Conference
  on Computer Vision and Pattern Recognition. pp. 9054--9063 (2021)

\bibitem{pumarola2020d}
Pumarola, A., Corona, E., Pons-Moll, G., Moreno-Noguer, F.: D-nerf: Neural
  radiance fields for dynamic scenes. In: Proceedings of the IEEE/CVF
  Conference on Computer Vision and Pattern Recognition. pp. 10318--10327
  (2021)

\bibitem{reddy2021tessetrack}
Reddy, N.D., Guigues, L., Pishchulin, L., Eledath, J., Narasimhan, S.G.:
  Tessetrack: End-to-end learnable multi-person articulated 3d pose tracking.
  In: Proceedings of the IEEE/CVF Conference on Computer Vision and Pattern
  Recognition. pp. 15190--15200 (2021)

\bibitem{rematas2021urf}
Rematas, K., Liu, A., Srinivasan, P.P., Barron, J.T., Tagliasacchi, A.,
  Funkhouser, T., Ferrari, V.: Urban radiance fields. In: Proceedings of the
  IEEE/CVF Conference on Computer Vision and Pattern Recognition. pp.
  12932--12942 (2022)

\bibitem{schmidt2015dart}
Schmidt, T., Newcombe, R., Fox, D.: Dart: dense articulated real-time tracking
  with consumer depth cameras. Autonomous Robots  \textbf{39}(3),  239--258
  (2015)

\bibitem{sigal2010humaneva}
Sigal, L., Balan, A.O., Black, M.J.: Humaneva: Synchronized video and motion
  capture dataset and baseline algorithm for evaluation of articulated human
  motion. International journal of computer vision  \textbf{87}(1),  4--27
  (2010)

\bibitem{srinivasan2021nerv}
Srinivasan, P.P., Deng, B., Zhang, X., Tancik, M., Mildenhall, B., Barron,
  J.T.: Nerv: Neural reflectance and visibility fields for relighting and view
  synthesis. In: Proceedings of the IEEE/CVF Conference on Computer Vision and
  Pattern Recognition. pp. 7495--7504 (2021)

\bibitem{starke2019neural}
Starke, S., Zhang, H., Komura, T., Saito, J.: Neural state machine for
  character-scene interactions. ACM Trans. Graph.  \textbf{38}(6),  209--1
  (2019)

\bibitem{stone2001blind}
Stone, J.V.: Blind source separation using temporal predictability. Neural
  computation  \textbf{13}(7),  1559--1574 (2001)

\bibitem{su2021anerf}
Su, S.Y., Yu, F., Zollhoefer, M., Rhodin, H.: A-nerf: Articulated neural
  radiance fields for learning human shape, appearance, and pose. In: NeurIPS
  (2021)

\bibitem{tancik2022blocknerf}
Tancik, M., Casser, V., Yan, X., Pradhan, S., Mildenhall, B., Srinivasan, P.P.,
  Barron, J.T., Kretzschmar, H.: Block-nerf: Scalable large scene neural view
  synthesis. In: Proceedings of the IEEE/CVF Conference on Computer Vision and
  Pattern Recognition. pp. 8248--8258 (2022)

\bibitem{TewariFTSLSMSSN20}
Tewari, A., Fried, O., Thies, J., Sitzmann, V., Lombardi, S., Sunkavalli, K.,
  Martin{-}Brualla, R., Simon, T., Saragih, J.M., Nie{\ss}ner, M., Pandey, R.,
  Fanello, S.R., Wetzstein, G., Zhu, J., Theobalt, C., Agrawala, M., Shechtman,
  E., Goldman, D.B., Zollh{\"{o}}fer, M.: State of the art on neural rendering.
  Comput. Graph. Forum  \textbf{39}(2),  701--727 (2020)

\bibitem{Tretschk_2021_ICCV}
Tretschk, E., Tewari, A., Golyanik, V., Zollh\"ofer, M., Lassner, C., Theobalt,
  C.: Non-rigid neural radiance fields: Reconstruction and novel view synthesis
  of a dynamic scene from monocular video. In: Proceedings of the IEEE/CVF
  International Conference on Computer Vision. pp. 12959--12970 (October 2021)

\bibitem{turki2021mega}
Turki, H., Ramanan, D., Satyanarayanan, M.: Mega-nerf: Scalable construction of
  large-scale nerfs for virtual fly-throughs. In: Proceedings of the IEEE/CVF
  Conference on Computer Vision and Pattern Recognition. pp. 12922--12931
  (2022)

\bibitem{vedula1999three}
Vedula, S., Baker, S., Rander, P., Collins, R., Kanade, T.: Three-dimensional
  scene flow. In: Proceedings of the Seventh IEEE International Conference on
  Computer Vision. vol.~2, pp. 722--729. IEEE (1999)

\bibitem{vogel2013piecewise}
Vogel, C., Schindler, K., Roth, S.: Piecewise rigid scene flow. In: Proceedings
  of the IEEE International Conference on Computer Vision. pp. 1377--1384
  (2013)

\bibitem{wang2021neural}
Wang, C., Eckart, B., Lucey, S., Gallo, O.: Neural trajectory fields for
  dynamic novel view synthesis. arXiv preprint arXiv:2105.05994  (2021)

\bibitem{xian2020space}
Xian, W., Huang, J.B., Kopf, J., Kim, C.: Space-time neural irradiance fields
  for free-viewpoint video. In: Proceedings of the IEEE/CVF Conference on
  Computer Vision and Pattern Recognition. pp. 9421--9431 (2021)

\bibitem{xiangli2021citynerf}
Xiangli, Y., Xu, L., Pan, X., Zhao, N., Rao, A., Theobalt, C., Dai, B., Lin,
  D.: Citynerf: Building nerf at city scale. arXiv preprint arXiv:2112.05504
  (2021)

\bibitem{xie2021neural}
Xie, Y., Takikawa, T., Saito, S., Litany, O., Yan, S., Khan, N., Tombari, F.,
  Tompkin, J., Sitzmann, V., Sridhar, S.: Neural fields in visual computing and
  beyond. In: Computer Graphics Forum. vol.~41, pp. 641--676. Wiley Online
  Library (2022)

\bibitem{yang2021learning}
Yang, B., Zhang, Y., Xu, Y., Li, Y., Zhou, H., Bao, H., Zhang, G., Cui, Z.:
  Learning object-compositional neural radiance field for editable scene
  rendering. In: Proceedings of the IEEE/CVF International Conference on
  Computer Vision. pp. 13779--13788 (2021)

\bibitem{yoon2020novel}
Yoon, J.S., Kim, K., Gallo, O., Park, H.S., Kautz, J.: Novel view synthesis of
  dynamic scenes with globally coherent depths from a monocular camera. In:
  Proceedings of the IEEE/CVF Conference on Computer Vision and Pattern
  Recognition. pp. 5336--5345 (2020)

\bibitem{zhai2021optical}
Zhai, M., Xiang, X., Lv, N., Kong, X.: Optical flow and scene flow estimation:
  A survey. Pattern Recognition  \textbf{114},  107861 (2021)

\bibitem{ZhangLYZZWZXY21}
Zhang, J., Liu, X., Ye, X., Zhao, F., Zhang, Y., Wu, M., Zhang, Y., Xu, L., Yu,
  J.: Editable free-viewpoint video using a layered neural representation.
  {ACM} Trans. Graph.  \textbf{40}(4),  149:1--149:18 (2021)

\end{thebibliography}

\appendix
\beginsupplement

\newpage
\noindent
\pdfbookmark[0]{Supplementary Material}{sup_mat}
{\Large\textbf{Supplementary Material} \vspace{1em}}

\section{Scope and Requirement Comparison}
We compare the settings in our paper with other similar works in \cref{tab:related-work}.
\begin{table}[h]
    \footnotesize
    \centering
    \caption{
    Comparison of methods for 3D motion estimation.
    \textit{Topologically varying scenes} mean that new surfaces may appear mid sequence; \textit{physical motion} means that methods estimate the motion of real 3D points, as opposed to canonical motion.
    }
    \begin{tabular}{cccc}
    \toprule
    Method & 
    \begin{tabular}{c}Topologically\\Varying Scenes\end{tabular}~~ &  
    \begin{tabular}{c}Physical\\Motion\end{tabular} & 
    \begin{tabular}{c}Precomputed\\Data-driven Prior\end{tabular} \\ \midrule
    D-NeRF \cite{pumarola2020d} & \xmark & \cmark & None \\
    Nerfies \cite{Park_2021_ICCV} & \xmark & \cmark & None  \\
    NR-NeRF \cite{Tretschk_2021_ICCV} & \xmark & \cmark & None \\
    DCT-NeRF \cite{wang2021neural} & \xmark & \cmark & Depth \& Optical Flow \\
    NSFF \cite{li2020neural} & \cmark & \cmark & Depth \& Optical Flow \\
    VideoNeRF \cite{xian2020space} & \cmark & \cmark & Depth \\
    NeRFlow \cite{du2021nerflow} & \cmark & \cmark & Optical Flow \\
    HyperNeRF \cite{park2021hypernerf} & \cmark & \xmark & None \\
    PREF (Ours) & \cmark & \cmark & None \\
    \bottomrule
    \end{tabular}
    \label{tab:related-work}
    \vspace{-1em}
\end{table}

\section{Implementation and Experimental Details}

\subsection{Architecture}
We use the same architecture as in NeRF for the space-time field: The positional encoding of the input location ($x$) and the timestamp $t$ is passed through 8 fully-connected ReLU layers. Each FC layer is with 256 channels. A skip connection is added to concatenate the input to the 5th layer’s activation. Note that different from NeRF, in our experiments, the feature vector from \nth{8} layer is not concatenated with the viewing directions. 
The predictor consists of 5 fully connected layers with a width of 128 and ReLU activations. The input of the predictor is the concatenation of three motion weights.
The motion field uses the same architecture as the space-time field, except the inputs are now location and motion embedding and the outputs are the motion vector.

\subsection{Dataset}
We select four representative clips from the Panoptic dataset \cite{joo2015panoptic}. The details of the four clips are as follows:
\begin{center}
\begin{tabular}{c@{\quad}c@{\quad}c@{\quad}c}
\toprule
     & full name & starting frame & end frame \\
     \midrule
    \textsc{Sports} & \texttt{161029\_sports1} & 3600 & 4000 \\
    \textsc{Tools} & \texttt{161029\_tools1} & 3700 & 4100 \\
    \textsc{Ian} & \texttt{160906\_ian5} & 600 & 1000 \\
    \textsc{Cello} & \texttt{171026\_cello3} & 220 & 620 \\
    \bottomrule
\end{tabular}
\end{center}
Full name indicates the name of the full sequence in the dataset. Starting frame and end frame are the frame index acquired with the official video-to-image conversion scripts.

\section{Additional Experimental Results}

\subsection{Validation on 2D Toy Example}
Fig. \ref{fig:toy} provides some quantitative results on a 2D toy dataset for validating the proposed predictability regularization.

\subsection{Video Results}

Readers can find a high-definition video, illustrating our methods and its results on a variety of dataset, at \href{http://pref.uiius.com}{pref.uiius.com}.
The video is encoded with H.264 codec, which can be downloaded here:
\url{https://www.codecguide.com/download\_kl.htm}. 

\subsection{Pose Tracking Visualization}
\begin{figure}[H]
    \centering
    \includegraphics[width=1.0\columnwidth,trim={0 0 0 26},clip]{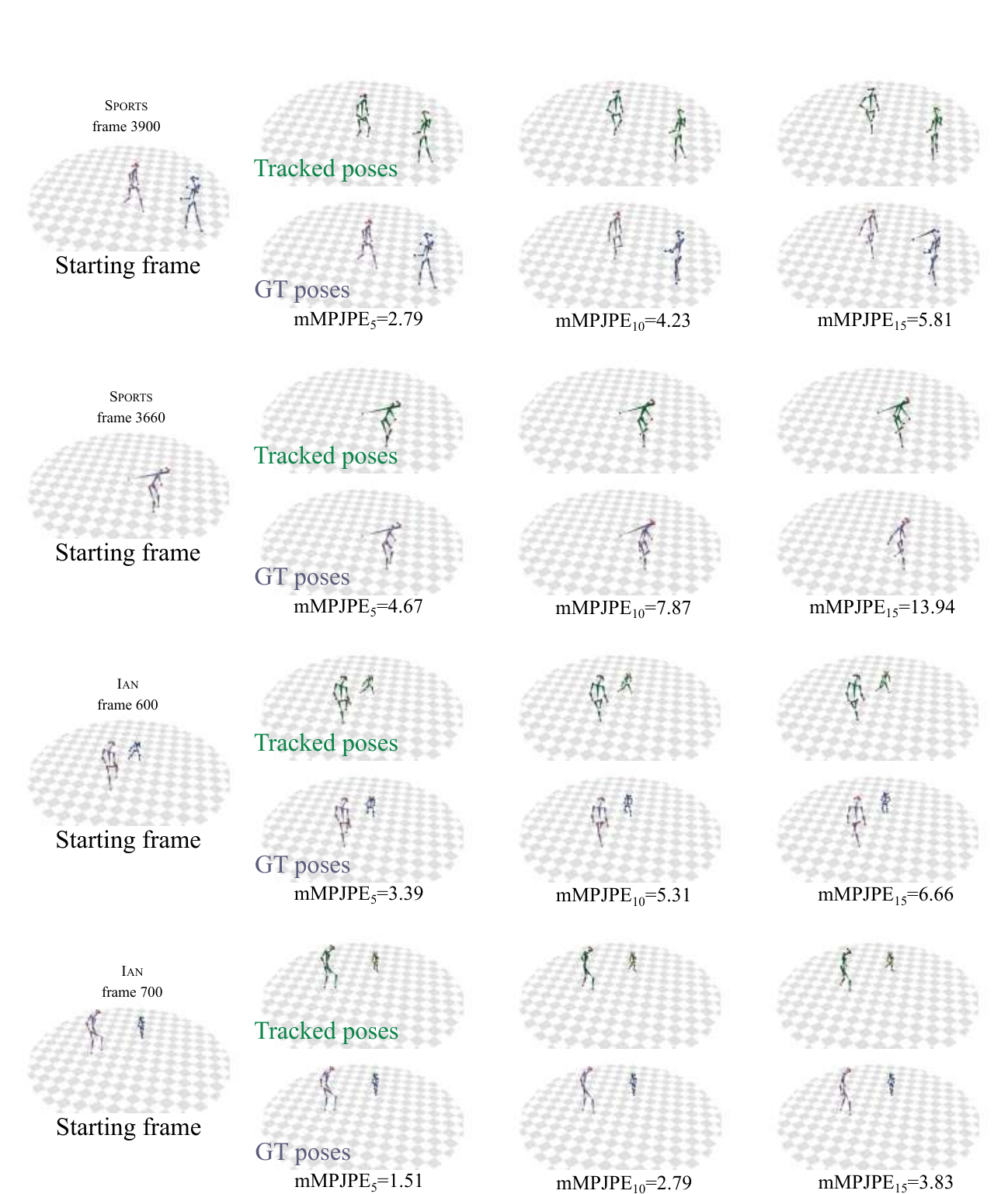}
    \caption{Visualization of motion tracking results on the Panoptic dataset.}
    \label{fig:mpjpe:sup}
\end{figure}

We provide additional visualizations of the tracked pose in \cref{fig:mpjpe:sup}. We can observe that some provided body joints are wrong in Panoptic, such as the top two rows of the \textsc{Sports} sequence. This is caused by the inaccurate 3D detector used by Panoptic as the 3D joints are not manually labeled.

\newpage

\begin{figure}[H]
    \centering
    \begin{animateinline}[controls,buttonsize=.7em,autoplay,loop,poster=last,width=1.\columnwidth]{1}
        \centering
        \multiframe{8}{i=0+1}{%
            \begin{tabular}{cc}
            (a) Input sequence & 
            (b) Ground-truth motion
            \\
            \includegraphics[width=.5\columnwidth]{rebuttal/fig/img\i} &
            \includegraphics[width=.5\columnwidth]{rebuttal/fig/gt\i}%
            \\
            (c) w/o predictability & 
            (d) with predictability \\
            Abs Err 0.01459 &
            Abs Err 0.00684 \\
            \includegraphics[width=.5\columnwidth]{rebuttal/fig/wo_pred_\i} &
            \includegraphics[width=.5\columnwidth]{rebuttal/fig/with_pred_\i}%
            \\
            \multicolumn{2}{p{\textwidth}}{\scriptsize{To see the animation, please view the document with compatible software, \eg,  \textit{Adobe Acrobat} or \textit{KDE Okular}; otherwise, the animation is also provided as a separate file attached to the submission.}}
            \end{tabular}
        }
    \end{animateinline}
    \caption{Ablation study on a toy example. A 2D neural motion field (pixel coordinate $uv$ as inputs) is studied. Two deformation patterns are applied alternatively. Abs Err is the mean absolute difference between the estimated and GT motion.
    }
    \label{fig:toy}
\end{figure}

\end{document}